\documentclass[letterpaper, 10 pt, conference]{ieeeconf}  

\IEEEoverridecommandlockouts                              

\usepackage{balance}
\usepackage{graphicx,cite}
\usepackage{bm}
\usepackage{amssymb}
\usepackage[ruled]{algorithm}
\usepackage{url}
\usepackage{color}

\newtheorem{theorem}{Theorem}

\newcommand{\V}[1]{\bm{#1} } 

\newcommand{\LOOE}{\epsilon_{\rm LOO}}

\newcommand{\nnzero}{K}

\newcommand{\noise}{n}

\newcommand{\be}{\begin{eqnarray}}
\newcommand{\ee}{\end{eqnarray}}
\newcommand{\ba}{\begin{array}}
	\newcommand{\ea}{\end{array}}

\newcommand{\subbe}{\begin{subequations}}
	\newcommand{\subee}{\end{subequations}}

\newcommand{\Hessian}{\partial^2_{\V{m}}}
\newcommand{\gradient}{\partial_{\V{m}}}

\title{\LARGE \bf
Approximate cross--validation formula for Bayesian linear regression*
}

\author{Yoshiyuki Kabashima$^{1}$, Tomoyuki Obuchi$^{2}$ and Makoto Uemura$^{3}$
\thanks{*This work was supported in part by JSPS KAKENHI Nos.\ 25120013 (YK), 26870185 (TO), and 25120007 (MU). The UC Berkeley SNDB is acknowledged for the permission to use the data set in section \ref{supernova}.  }
\thanks{$^{1}$Yoshiyuki Kabashima is with Department of Mathematical and Computing Science, 
			Tokyo Institute of Technology, Yokohama 226-8502, Japan
        {\tt\small kaba@c.titech.ac.jp}}%
\thanks{$^{2}$Tomoyuki Obuchi is with Department of Mathematical and Computing Science, Tokyo Institute of Technology, Yokohama 226-8502, Japan
        {\tt\small obuchi@c.titech.ac.jp}}%
\thanks{$^{3}$Makoto Uemura is with Hiroshima Astrophysical Science Center, 
			Hiroshima University, 
			Kagamiyama 1-3-1, Higashi-Hiroshima, 739-8526, Japan
	{\tt\small uemuram@hiroshima-u.ac.jp}}%
}

\begin{document}

\maketitle
\thispagestyle{empty}
\pagestyle{empty}

\begin{abstract}

		Cross--validation (CV) is a technique for evaluating the ability of statistical models/learning systems based
		on a given data set. Despite its wide applicability, the rather
		heavy computational cost can prevent its use as the system size
		grows. To resolve this difficulty in the case of Bayesian
		linear regression, we develop a formula for 
		evaluating the leave-one-out CV error approximately without
		actually performing CV. 
		The usefulness of the developed formula is tested by
		statistical mechanical analysis for a synthetic model. This is  
		confirmed by application to a real-world supernova data set as well.  

\end{abstract}

\section{Introduction}
	Consider carrying out linear regression analysis for a data set 
	$D^M=\{(\V{x}_{\mu},y_\mu)\}_{\mu=1}^M$, where $\V{x}_{\mu} =(x_{i\mu})\in \mathbb{R}^N$ and $y_\mu \in \mathbb{R}$. 
	The linear regression model assumes that the relationship between $y_\mu$ and $\V{x}_\mu$ is linear, 
	which indicates that the model takes the form 
	\begin{eqnarray}
	y_\mu = \V{x}_\mu^{\top} \V{w} + \noise_\mu, \quad \mu =1,2,\ldots, M
	\label{linear_regression}
	\end{eqnarray}
	using the parameter vector $\V{w}=(w_i)\in \mathbb{R}^N$. Here, $\top$ represents the 
	matrix-vector transpose operation, $\noise_\mu$ indicates noise and 
	the intercept term is omitted for simplicity. 
	
	The parameter vector $\V{w}$ can be determined uniquely by the least square method 
	if $M \ge N$. 
	However, in some contexts such as compressed sensing and certain kinds of 
	high-dimensional data analysis, one needs to infer $\V{w}$ even when $M < N$. The Bayesian framework offers a useful strategy for coping with such demands. 
	For this, we introduce a {\em sparse prior} probability distribution $P(\V{w}|\rho)=\prod_{i=1}^N\phi(w_i|\rho)$, 
	where $\phi(w_i|\rho)=(1-\rho)\delta(w_i)  +\rho f(w_i)$, $0\le \rho \le 1$ and 
	$f(w)$ is a density function that does not have finite mass at $w=0$,  
	which effectively suppresses the $\V{w}$ degree of freedom. 
	This {\em formally} yields the posterior distribution
	\begin{eqnarray}
	P(\V{w}|D^M;\beta, \rho)=
	\frac{e^{-\beta {\rm RSS}(\V{w}|D^M)}\prod_{i=1}^N\phi(w_i|\rho)}{Z(D^M;\beta,\rho)},  
	\label{posterior}
	\end{eqnarray}
	where $\beta^{-1}$ corresponds to the variance of noise, 
		\begin{eqnarray}
	Z(D^M;\beta,\rho)=\int e^{-\beta {\rm RSS}(\V{w}|D^M)}\prod_{i=1}^N\phi(w_i|\rho)d\V{w}, 
	\label{partition}
	\end{eqnarray}
	and 
	\begin{eqnarray}
	{\rm RSS}(\V{w}|D^M) =\frac{1}{2} \sum_{\mu=1}^M \left (y_\mu -\V{x}^{\top}_\mu \V{w} \right )^2
	\label{RSS}
	\end{eqnarray}
	means the residual sum of squares (RSS).
	Based on (\ref{posterior}), one can set the posterior mean 
	\begin{eqnarray}
	\left \langle \V{w} \right \rangle = \int \V{w} P(\V{w}|D^M;\beta, \rho) d \V{w}
	\label{postmean}
	\end{eqnarray}
	as a reasonable estimator of the parameter vector $\V{w}$
	since (\ref{postmean}) considerably reduces (\ref{RSS}) when the 
	hyper-parameters $\beta$ and $\rho$ are appropriately tuned. 
	
	However, there are two obstacles to making this procedure practical. 
	The first one is the computational difficulty of evaluating (\ref{postmean}). 
	For this, one of the current authors has
	developed an approximation method \cite{KabashimaVehekapera2014}, which can be utilized for systems of reasonable sizes. 
	The other issue, which we would like to address here, is how to determine $\beta$ and $\rho$. 
	If $D^M$ is actually generated by the process of (\ref{linear_regression})
	using a true parameter vector $\V{w}^0$ that follows the sparse prior $P(\V{w}|\rho)$ 
	and Gaussian noises, then maximization of the marginal likelihood
	$P(D^M;\beta,\rho)=\int P(D^M|\V{w},\beta)P(\V{w}|\rho) d\V{w} = 
	Z(D^M;\beta,\rho)/(2 \pi \beta^{-1} )^{M/2}$ would be the most rational approach. 
	Unfortunately, in practice, there are many situations where we cannot expect these assumptions to hold. 
	In such cases, especially when the objective of the regression is to maximize the prediction ability 
	for novel samples, minimizing the {\em cross--validation (CV) error} is a dominant alternative. 
	
	CV is a general framework for evaluating the prediction ability of statistical models/learning systems
	based on a given data set. Despite its wide applicability, the heavy computation cost can make
	it difficult to use. The purpose of this paper is to show that, in the case of the Bayesian linear regression, employment of 
	the approximation method of \cite{KabashimaVehekapera2014} naturally enables us 
	to evaluate, approximately, the leave-one-out (LOO) CV error without actually performing CV, which considerably reduces the computational cost. 
	The usefulness will also be tested by applications to a synthetic problem and a real world data analysis. 
	
	\section{Related work}
	There have been several studies of 
	CV for linear regression.  
	The analytical formula for evaluating the LOO CV error (LOOE) exactly, without actually performing 
	CV, is widely known for standard linear regression and ridge regression \cite{exactCV}. 
	This formula was extended to the case in which linear constraints are present \cite{Tarpey2000}. 
	An alternative measure, which has a property similar to that of LOOE and 
	can be evaluated at a lower computational cost, was proposed as the ``generalized cross--validation'' 
	in \cite{Golub1979} for regularized linear regression.  
	Two types of LOOE approximation formulas for LASSO were recently provided 
	in \cite{ObuchiKabashima2016}. In contrast to these, our aim here 
	is to develop a computationally feasible approximate formula to 
	evaluate LOOE in the Bayesian formalism 
	in which sparse (singular) priors can be employed.   
	A similar attempt has been made for feedforward neural networks in \cite{OpperWinther1996}.

	\section{Cross--validation in Bayesian linear regression}
	\subsection{Expectation consistent (EC) approximation}
	As the basis of our study, we briefly mention the approximate 
	inference scheme for Bayesian linear regression developed 
	in \cite{KabashimaVehekapera2014} and summarized in the following two theorems.  
	\begin{theorem}
		\label{theorem1}
		Let us define {\em Gibbs free energy} as
		\begin{eqnarray}
		&&\Phi(\V{m}|D^M; \beta,\rho )=\mathop{\rm extr}_{\V{h}}\left \{
		\V{h} \cdot \V{m} \right . \cr
		&&  \left . -\ln \left [
		\int \!
		\left (
		\! e^{-\beta {\rm RSS}(\V{w}|D^M)} \!\prod_{i=1}^N \! \phi(w_i|\rho) \! \right )
		e^{\V{h} \cdot \V{w}}d \V{w} 
		\right ]
		\right \},
		\label{Gibbs}
		\end{eqnarray}
		where $\mathop{\rm extr}_{u}\left \{ f(u) \right \}$ generally means the 
		extremization of a function $f(u)$ with respect to $u$. 
		The function $\Phi(\V{m}|D^M; \beta,\rho )$ is a downward-convex function of 
		$\V{m}=(m_i)\in \mathbb{R}^N$ and its unique minimizer accords to (\ref{postmean}). 
		
	\end{theorem}
	\begin{theorem}
		To characterize the property of the $N \times M $ data matrix $X=(x_{i\mu})$, 
		we introduce the function 
		\begin{eqnarray}
		G(x)&=&\mathop{\rm extr}_{\Lambda} 
		\left \{-\frac{1}{2 N} \ln \det \left (\Lambda-XX^{\top} \right )+\frac{\Lambda x}{2} \right \} \cr
		&& -\frac{1}{2}\ln x-\frac{1}{2}. 
		\label{Gfunc}
		\end{eqnarray}
		The expectation consistent (EC) approximation \cite{OpperWinther2001}, requiring matching of 
		the first and 
		the macroscopic second moments, approximately evaluates (\ref{Gibbs}) as 
		\begin{eqnarray}
		&&\Phi(\V{m}|D^M; \beta, \rho )  \simeq \Phi_{\rm EC} (\V{m}|D^M; \beta, \rho )\cr
		&& \equiv \mathop{\rm extr}_{Q,E,\V{h}} \! \left \{ \!
		\beta {\rm RSS}(\V{m}|D^M) \!- \!NG\left ( \!- \!\beta (Q-q) \right )  \!- \!\frac{NEQ}{2} \right .  \cr
		&& \left .
		+ \V{h} \cdot \V{m}
		- \sum_{i=1}^N \ln \left [\int  \phi(w_i|\rho)   e^{-\frac{E}{2}w_i^2+h_i w_i} d w_i\right ]\right \}\cr
		&&+const,
		\label{EC}
		\end{eqnarray}
		for large systems, where $q=N^{-1} |\V{m}|^2$. 
	\end{theorem}
	
	The detailed derivations of the two theorems are provided in \cite{KabashimaVehekapera2014}. 
	These theorems indicate that the Bayesian estimator (\ref{postmean}) of the parameter vector 
	can be evaluated approximately by minimizing the EC free energy (\ref{EC}). Unfortunately, 
	the minimization is non-trivial to perform, and various methods have been proposed 
	for accomplishing this task \cite{Winther,Opper,Rangan}. 
	However, when the dimensionality $N$ is ``moderately large,'' Newton's method works pretty well, 
	and we here regard it as a default solver. More concretely, 
	one can obtain a solution of the EC approximation 
	as a convergent point of the following discrete dynamics counted by 
	$t=1,2,\ldots$: 
	\begin{eqnarray}
	\V{m}^{t+1}=\V{m}^{t}-\left (\Hessian \Phi_{\rm EC}^t\right )^{-1} \gradient \Phi_{\rm EC}^t,  
	\label{Newton}
	\end{eqnarray}
	where
	\begin{eqnarray}
	\gradient \Phi_{\rm EC}^t =-\beta X(\V{y}-X^{\top}\V{m}^t)
	-E^t \V{m}^t +\V{h}^{t}, 
	\label{Newton_grad}
	\end{eqnarray}
	\begin{eqnarray}
	\Hessian \Phi_{\rm EC}^t=\beta X X^\top+\left (\left [
	\frac{1}{M_i^t-(m_i^t)^2 } -E^t \right ]\delta_{ij} \right ) , 
	\label{chi}
	\end{eqnarray}
	and $\V{y}=(y_\mu)$. 
	Here, $\V{h}^t=(h_i^t)$ and $E^t$ stand for
	the extremized values of $\V{h}=(h_i)$ and $E$ for $\V{m}=\V{m}^t=(m_i^t)$ 
	at the right hand side of (\ref{EC}), respectively, and 
	$M_i^t=-2 (\partial/\partial E^t) \ln \left [\int  \phi(w_i|\rho)   e^{-\frac{E^t}{2}w_i^2+h_i^t w_i} d w_i\right ]$. 
	For simplicity, terms of $O(N^{-1})$ are omitted in (\ref{chi}). 
	
	The most time-consuming part in the above calculation 
	is the matrix inversion operation required in 
	(\ref{chi}). Therefore, the computation cost per update scales roughly as $O(N^3)$.  
	
	\subsection{Cross--validation (CV)}
	The EC approximation offers an approximate estimate of (\ref{postmean}) for a given pair of 
	hyper-parameters $\beta$ and $\rho$. However, in many situations, 
	their correct values are not provided in advance and have to be determined from $D^M$. 
	The minimized value of (\ref{Gibbs}) generally corresponds to the negative 
	logarithm of the partition function $Z(D^M;\beta,\rho)$. 
	As $Z(D^M;\beta,\rho)$ is proportional to the marginal likelihood function
	$P(D^M;\beta,\rho)$, determining $\beta$ and $\rho$ such that they 
	minimize $\mathop{\rm min}_{\V{m}} \left \{\Phi_{\rm EC} (\V{m}|D^M; \beta, \rho ) \right \}$
	is a reasonable strategy in terms of the maximum likelihood principle. 
	Indeed, Ref.~\cite{Krzakala2012} reports that a similar strategy exhibits an excellent 
	inference performance for signal recovery of compressed sensing. 
	Nevertheless, this is not necessarily the case when the correct posterior cannot 
	be expressed by the assumed model class, which would be more common in practice. 
	
	In such cases, under the assumption that the data $(\V{x}_\mu, y_\mu)$ 
	of $\mu =1,2,\ldots, M$ are generated 
	independently from an identical distribution, 
	maximizing the prediction ability for novel samples,
	is an alternative guideline for determining the hyper-parameters. 
	When no extra data is available other than $D^M$, CV  
	is a widely used method for estimating the prediction ability. 
	In the current problem, the standard $k$-fold CV 
	is carried out as follows:
	the data set $D^M$ is first divided into $k$ non-overlapping subsets, 
	and the data of each subset are predicted using the estimator of 
	(\ref{postmean}) that is determined from the data of the remaining $k-1$ subsets. 
	Then, the CV error, which measures the prediction ability, is computed as the average 
	RSS for the prediction over the $k$ choices of the test subsets. This 
	means that performing the $k$-fold CV enforces us 
	to carry out the minimization of (\ref{EC}) $k$ times. 
	Furthermore, we need to repeat the CV procedure many times for optimizing $\beta$ and $\rho$. 
	Although the minimization of $(\ref{EC})$ for the $k$ choices of test subsets 
	is easily parallelized, the heavy computational cost could make the CV-based 
	hyper-parameter determination practically infeasible. 
	
	\subsection{Leave-one-out (LOO) CV and its approximate formula}
	To reduce the computational cost, let us develop an
	approximate formula that estimates the CV error without performing CV.
	For this, we take particular note of 
	LOO CV, 
	which corresponds to $k=M$. As the size of each test subset of LOO CV is 
	only unity, the differences between the estimators of an LOO and the full data sets is expected to be small. 
	This enables us to evaluate the estimator of the LOO data set from that of the full data set 
	in a perturbative manner, which yields a semi-analytic formula to evaluate the CV error based on 
	the result of the EC approximation for the full data set $D^M$. 
	
	The fixed--point condition of (\ref{Newton}), which can be read as
	\begin{eqnarray}
	&&m_i=f\left (h_i; E \right ),
	\label{fixed_point1} \\
	&& h_i = \beta \sum_{\mu=1}^M x_{i\mu} (y_\mu - \sum_{j=1}^N x_{j\mu} m_j)+E m_i,
	\label{fixed_point2}
	\end{eqnarray}
	constitutes the basis for implementing the above idea. 
	Here, we have defined a function $f(h;E)$ by 
	$f(h;E)=(\partial/\partial h)\ln \left [\int \phi(w|\rho)e^{-\frac{E}{2}w^2+h w} dw \right ]$. 
	Let us denote the solution of (\ref{fixed_point1}) and (\ref{fixed_point2}) for the ``$\mu$-th LOO system,''  which is defined by leaving
	the $\mu$-th data $(\V{x}_\mu, y_\mu)$ out from the full data set, 
	as $m_{i \to \mu}$ and $h_{i\to \mu}$.  
	Since the contribution of the $\mu$-th data $(\V{x}_\mu, y_\mu)$ to 
	(\ref{fixed_point1}) and (\ref{fixed_point2}) is supposed to be small, the relation
	$h_i \simeq h_{i\to \mu} +\Delta h_{\mu \to i}$ holds, where
	\begin{eqnarray}
	\Delta h_{\mu \to i} &\equiv& \beta x_{i\mu} \left (y_\mu -\sum_{ j=1}^N x_{j \mu} m_{j\to \mu} \right ) \cr
	&\simeq& \beta x_{i\mu} \left (y_\mu -\sum_{ j=1}^N x_{j \mu} m_{j} \right )
	\label{cavity_bias}
	\end{eqnarray}
	is regarded as small. 
	This indicates that the relation 
	\begin{eqnarray}
	m_i \simeq m_{i\to \mu} + \sum_{j=1}^N c_{ij}^{\backslash \mu} \Delta h_{\mu \to j}, 
	\label{full_cavity}
	\end{eqnarray}
	holds between $m_i$ and $m_{i \to \mu}$, where $c_{ij}^{\backslash \mu}$ represents the rate of change of $m_{i \to \mu}$ when 
	$h_{j\to \mu}$ is slightly changed. 
	
	To evaluate $c_{ij}^{\backslash \mu}$, we 
	add the ``external fields'' $\theta_j$ to $h_{j \to \mu}$ for $\forall{j} \in \{1,2,\ldots, N\}$ 
	in (\ref{fixed_point1}) of the $\mu$-th LOO system, 
	and take the partial derivative with respect to $\theta_j$  for $\forall{j}$ at $(\theta_j)=\V{0}$ 
	\cite{OpperWinther2001}. 
	This yields a set of $N^2$ coupled equations 
	\begin{eqnarray}
	c_{ij}^{\backslash \mu} \!= \! (M_i \!- \! m_i^2) \!\left ( \! \delta_{ij} \!-\!\beta \sum_{k=1}^N \sum_{\nu \ne \mu} x_{i\nu}x_{k \nu}
	c_{kj}^{\backslash \mu} \!+\! E c_{ij}^{\backslash \mu} \! \right ), 
	\label{chi_eq}
	\end{eqnarray}
	where we used the relation $\partial f(h_{i\to \mu};E)/\partial h_{i\to \mu}
	= M_{i\to \mu}-m_{i\to \mu}^2 \simeq M_i-m_i^2$. 
	The solution of (\ref{chi_eq}) is given as
	\begin{eqnarray}
	(c_{ij}^{\backslash \mu}) =\left (\Hessian \Phi_{\rm EC}-\beta \V{x}_\mu \V{x}_\mu^\top \right )^{-1}
	\label{chi_sol}
	\end{eqnarray}
	in the matrix expression, where $\Hessian \Phi_{\rm EC}$ stands for the Hessian of the EC free energy (\ref{EC})
	at the full estimator $\V{m}=(m_i)$. 
	
	The expressions (\ref{full_cavity}) and (\ref{chi_sol}) can be employed to evaluate 
	the CV error of the $\mu$-th LOO estimator $(m_{i\to \mu})$ on the $\mu$-th data $(\V{x}_\mu, y_\mu)$ 
	using the full estimator $\V{m}=(m_i)$. 
	For this, we evaluate the residual of the $\mu$-th LOO estimator
	on $(\V{x}_\mu, y_\mu)$  utilizing (\ref{cavity_bias}) and (\ref{full_cavity}) as
	\begin{eqnarray}
	&&y_\mu \!-\!\sum_{i=1}^N x_{i \mu }m_{i\to \mu} 
	\!\simeq \! y_\mu \!-\! \sum_{i=1}^N x_{i \mu } \!
	\left (m_{i} \!- \!\sum_{j=1}^N c_{ij}^{\backslash \mu} \! \Delta h_{\mu \to j} \! \right ) \cr
	&&= \! \left (1+\beta \sum_{i,j} x_{i\mu}x_{j\mu}c_{ij}^{\backslash \mu} \right ) 
	\left (y_\mu -\sum_{i=1}^N x_{i \mu }m_i \right ) \cr
	&&=\! \left (1 \!- \!\beta \V{x}_\mu^\top (\Hessian \Phi_{\rm EC})^{-1}  \!  \V{x}_\mu \right )^{-1} 
	\! \left (y_\mu \! \!- \! \!\sum_{i=1}^N x_{i \mu }m_i \! \right ), \phantom{abcd}
	\end{eqnarray}
	where we used the Sherman-Morrison formula for the matrix inversion to simplify the expression. 
	This means that LOOE can be evaluated approximately as
	\begin{eqnarray}
	\LOOE &\equiv & \frac{1}{2M}\sum_{\mu=1}^M \left (y_\mu -\sum_{i=1}^N x_{i \mu }m_{i\to \mu} \right )^2\cr
	&\simeq & \frac{1}{2M}\sum_{\mu=1}^M \frac{\left (y_\mu -\sum_{i=1}^N x_{i \mu }m_{i} \right )^2}
	{\left (1-\beta \V{x}_\mu^\top (\Hessian \Phi_{\rm EC})^{-1}  \V{x}_\mu  \right )^2}. 
	\label{CVformula}
	\end{eqnarray}

	To evaluate LOOE literally, we need to solve the minimization problem 
	of (\ref{EC}) for $M$ LOO systems, which requires 
	$O(M N^3)$ computational costs, even if the necessary number of 
	iterations for convergence of (\ref{Newton}) is $O(1)$.   
	On the other hand, the approximate LOOE formula of (\ref{EC})
	is expressed by using the solution of the full system only. 
	In (\ref{Newton}), we need to evaluate the matrix inversion of the Hessian 
	$\Hessian \Phi_{\rm EC}$, which requires $O(N^3)$ computations. 
	However, this is already computed when performing (\ref{Newton})
	to obtain the full solution, and does not require any extra computation cost. 
	This means that the approximate formula accelerates the computation of 
	LOOE by a factor $M$. 
	
	\section{Numerical validation}
	\subsection{Synthetic model}
	We first examine the usefulness of the developed CV formula by applying it to a synthetic model in 
	which the vector $\V{y}$ is generated in the manner of (\ref{linear_regression})
	on the basis of a true sparse vector $\V{w}^0=(w_i^0)$. 
	For analytical tractability, we assume that $X$ is a simple random matrix whose entries are sampled 
	independently from ${\cal N}(0, N^{-1})$, and that $w_i^0$ 
	$(i=1,2,\ldots,N)$ and $\noise_\mu$ $(\mu=1,2,\ldots,M$) are also independently generated from 
	$\phi(w|\rho_0,\sigma_{w0}^2)=(1-\rho_0)\delta(w)+\rho_0 {\cal N}(0, \sigma_{w0}^2)$ and ${\cal N}(0, \sigma_{\noise 0}^2)$, 	respectively. Under these assumptions, 
	the replica method of statistical mechanics makes it possible to assess theoretically the typical values of various 	
	macroscopic quantities as $N, M \to \infty$ while keeping $\alpha =M/N $ finite~\cite{KabashimaVehekapera2014,Nakanishi:15}. 
	We performed the theoretical assessment under the so-called {\em replica symmetric} assumption.
	
	In the experiment, the system size was set to $N=1000$, and the system parameters were fixed to $\rho_0=0.1$, $\alpha=0.5$, 
	$\sigma_{w0}^2=10$, and $\sigma_{\noise 0}^2=0.1$. 
	We estimated $\V{w}^0$ following the Bayesian linear regression 
	utilizing the EC approximation (\ref{Newton}). 
	The prediction ability of the Bayesian estimator is optimized when the correct 
	sparse prior $\phi(w|\rho_0,\sigma_{w0}^2)$ and $\beta=\sigma_{\noise 0}^{-2}$ are employed. 
	However, such information is not available in many practical situations. 
	To examine whether the approximate CV formula (\ref{CVformula}) 
	offers a clue to optimize the prediction ability or not, 
	we set the prior as $\phi(w|\rho, \sigma_w^2)$ and 
	evaluated (\ref{CVformula}) changing hyper-parameters $\beta$, $\rho$, and $\sigma_w^2$. 
	
	
		\begin{figure}[t]
		\begin{center}
			\includegraphics[width=\columnwidth]{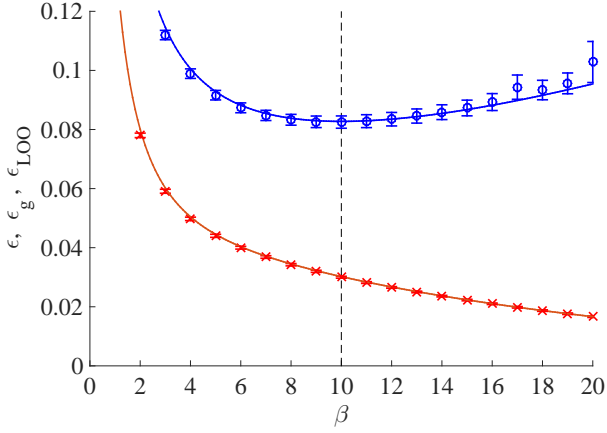}
		\end{center}
		\caption{\label{fig1}
			Comparison between theory and experiment for 
			$\alpha=0.5$, $\sigma_{w0}^2=10$, and $\sigma_{\noise 0}^2=0.1$. 
			For details, see the main text. 
			The experimental data are evaluated from $N_{\rm sample}=30$ samples of $N=1000$, 
			and the error bars represent the standard deviations among those samples
			divided by $\sqrt{N_{\rm sample}-1}$.
		}
	\end{figure}
	
	
	Figure \ref{fig1}  shows the result obtained by varying 
	$\beta$ while setting the remaining two hyper-parameters to 
	the correct values $\rho=\rho_0$ and $\sigma_w^2=\sigma_{w0}^2$. 
	The curves represent the theoretical estimate of 
	the typical value of achieved RSS per data 
	$\epsilon =(1/2M) \sum_{\mu=1}^M (y_\mu - \V{x}_\mu^\top \V{m})^2$
	(red curve) and that of the prediction error 
	for a new ($\mu+1$-st) data $\epsilon_{\rm g} =(1/2)  (y_{\mu+1} - \V{x}_{\mu+1}^\top \V{m})^2$
	(blue curve). 
	Although $\epsilon$ decreases monotonically as $\beta$ grows, 
	which indicates overfitting to the given data $D^M$, 
	$\epsilon_{\rm g}$ is minimized at the correct value 
	$\beta=\sigma_{n0}^{-2}=10$.
	The symbols stand for experimentally evaluated  
	$\epsilon$ (red crosses) and 
	LOOE assessed by the approximate formula (\ref{CVformula})
	$\epsilon_{\rm LOO}$ (blue circles), which were obtained from 30 experimental samples.  
	These samples exhibit good consistency with the theoretical estimates 
	of $\epsilon$ and $\epsilon_{\rm g}$.  
	Especially, the consistency between $\epsilon_{\rm g}$ and $\epsilon_{\rm LOO}$ 
	means that (\ref{CVformula}) offers a reliable estimate of 
	the prediction error based on a given data set. 
	This would be useful for determining the hyper-parameters to optimize the 
	prediction ability. 
	Similar results are obtained when $\rho$ and $\sigma_w^2$ are varied.

	\subsection{Type Ia Supernova data set}
	\label{supernova}
	\begin{table}[t]
		\small
		\begin{center}
			\begin{tabular*}{\columnwidth}{@{\extracolsep{\fill}}l ||c|c|c|c |c |c } 
				$\nnzero$ & 1 & 2 & 3 & 4 & 5 & 6\\
				\hline
				\hline 
				Approx. & 0.0328& 0.0235 &  0.0219 & 0.0218& 0.0220 & 0.0222\\
				Literal & 0.0327 & 0.0231 & 0.0220 & 0.0218 & 0.0218 & 0.0219 \\
				\hline
				RSS & 0.0312 & 0.0178 & 0.0163 & 0.0156 & 0.0152 & 0.0150
			\end{tabular*}
			\vspace*{2mm}
			\caption{\label{tab:supernova_LOOE}
				LOOEs obtained for $\nnzero=1$--$6$ for type Ia supernova data set. 
			}
		\end{center}
	\end{table}
	
	We also applied our methodology to a data set from the SuperNova DataBase (SNDB) provided by the Berkeley Supernova Ia program \cite{Berkeley,Silverman:12}.
	Screening based on a certain criteria 
	yields a reduced data set of $M = 78$ and $N = 276$ \cite{Uemura:15}.
	The purpose of the data analysis is to construct a formula to estimate accurately the absolute magnitude at the maximum of type 
	Ia supernovae by linear regression. 
	To estimate this quantity accurately is particularly important 
	in modern astronomy because it directly influences the measurement of 
	long distances in the universe.

	\begin{figure}[t]
		\begin{center}
			\includegraphics[width=\columnwidth]{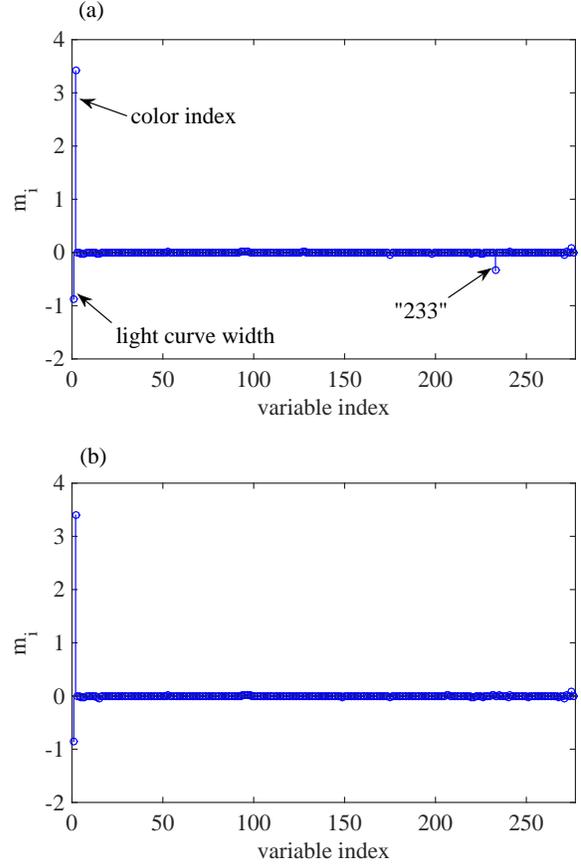}
		\end{center}
		\vspace*{-.8cm}
		\caption{\label{fig2}
			(a): Estimated parameter vector $\V{m}=(m_i)$ $(i=1,2,\ldots,276)$ by Bayesian sparse linear regression
			for $K=4$. (b): The same as in (a) when $m_{233}$ is forced to zero. 
			The results are not very sensitive to the choice of $K$; even if $K$ is set to $3$ or $5$ almost identical results are obtained. 
		}
	\end{figure}
	
	Following the conventional treatment of linear regression,
	we preprocessed both the absolute magnitude at the maximum (dependent variable) and the $276$ 
	candidates of explanatory variables, which are composed of processed spectral data, to have zero means.
	We applied the EC approximation of the Bayesian sparse linear regression to the preprocessed data set. 
	As the sparse prior, we employed the Bernoulli--Uniform distribution  
	$\phi_{\rm BU}(w|\rho) = (1-\rho)\delta(w)+\rho$ since we have no prior 
	knowledge about the distribution of the non-zero components. 
	The hyper-parameter $\rho$ was tuned to make the expected number of 
	non-zero components in the {\em posterior} distribution correspond to the controlled value $K(=1,2,\ldots)$. 
	For each $K$, the other hyper-parameter $\beta$ was determined so that
	the approximate LOOE (\ref{CVformula}) was minimized. 
	
	\balance
	
	To examine the validity of the approximate estimate, we
	also literally carried out LOO CV, utilizing the same values of 
	hyper-parameters as those of the approximate method. 
	Table I summarizes LOOEs evaluated for $K=1,2,\ldots,6$. 
	The values in the top row were obtained by the approximate CV formula of (\ref{CVformula})
	while those in the middle row were evaluated by literally performing LOO CV. 
	The values are reasonably consistent with each other, 
	which validates the usage of (\ref{CVformula}). 
	The bottom row contains the achieved RSS per data for reference.

	Table \ref{tab:supernova_LOOE} indicates that 
	LOOEs are minimized around $K=4$ while the achieved RSS per data 
	decreases monotonically as $K$ grows, which makes it possible to choose the optimal hyper-parameter as $K=4$ by monitoring the approximate 
	LOOE evaluated by (\ref{CVformula}). 
	Figure 2 (a) shows the stem plot of the estimator
	$\V{m}=(m_i)$ obtained for this optimal choice. 
	This indicates that most components except for the variables ``1'' and ``2,'' which stand for {\em light curve width} and {\em color index}, respectively, 
	are negligibly small. That said, variable ``233'' has the third largest amplitude, and may not be negligible. 
	In fact, this variable is defined by normalizing the spectral intensity of wavelength 6631~\AA~by the ``continuum level'' (roughly speaking, a locally smoothed intensity around the wave length), and hardly varies in the data set. This suggests that the possibly non-negligible amplitude 
	may be due to an accidental statistical fluctuation. 
	Indeed, analyzing the same data set enforcing $m_{233}=0$ produces 
	almost the same profile as in Fig.~2 (a) for other components (Fig.~2 (b)), 
	which implies that $m_{233}$ does not play an important role in the regression. 
	This may indicate that only the {\em light curve width} and {\em color index} are relevant for estimating the absolute magnitude at the maximum of type 
	Ia supernovae, which is consistent with the conclusions of earlier studies~\cite{Uemura:15,ObuchiKabashimaEUSIPCO2016}.

	\section{Summary and discussion}
	In summary, we developed a semi-analytic formula to evaluate the CV error approximately for Bayesian linear regression. 
	When the Hessian of the EC free energy is available, 
	the formula makes it possible to evaluate approximately the leave-one-out 
	CV error (LOOE) from the full estimator without actually performing 
	CV. The usefulness of the developed formula was tested and 
	confirmed by applications to a synthetic model and a real world data set of
	type Ia supernovae. 
	
	We make two observations: 
	First, although we employed the EC approximation, 
	similar formulas can also be developed in other approximation frameworks
	such as the naive mean field and the Bethe approximations. 
	Second, the standard linear regression, ridge regression, and LASSO can be formulated 
	as the maximum a posteriori estimator in the Bayesian framework. 
Techniques similar to that developed in the current paper 
can reproduce their existing formulas of LOOE by 
introducing an appropriate $\beta$--dependent prior and letting $\beta \to \infty$.  
	In this sense, the developed formula (\ref{CVformula}) can be regarded
	as a generalization of the existing formulas. 
	

\begin{thebibliography}{10}
\providecommand{\url}[1]{#1}
\csname url@samestyle\endcsname
\providecommand{\newblock}{\relax}
\providecommand{\bibinfo}[2]{#2}
\providecommand{\BIBentrySTDinterwordspacing}{\spaceskip=0pt\relax}
\providecommand{\BIBentryALTinterwordstretchfactor}{4}
\providecommand{\BIBentryALTinterwordspacing}{\spaceskip=\fontdimen2\font plus
\BIBentryALTinterwordstretchfactor\fontdimen3\font minus
  \fontdimen4\font\relax}
\providecommand{\BIBforeignlanguage}[2]{{%
\expandafter\ifx\csname l@#1\endcsname\relax
\typeout{** WARNING: IEEEtran.bst: No hyphenation pattern has been}%
\typeout{** loaded for the language `#1'. Using the pattern for}%
\typeout{** the default language instead.}%
\else
\language=\csname l@#1\endcsname
\fi
#2}}
\providecommand{\BIBdecl}{\relax}
\BIBdecl

\bibitem{KabashimaVehekapera2014}
{Y. Kabashima and M. Vehkaper\"{a}}, ``Signal recovery using expectation
  consistent approximation for linear observations,'' in \emph{ISIT2014
  Proceedings}, 2014, pp. 226--230.

\bibitem{exactCV}
{D. M. Allen}, ``The relationship between variable selection and data
  augmentation and a method for prediction,'' \emph{Technometrics}, pp.
  125--127, 1974.

\bibitem{Tarpey2000}
{T. Tarpey}, ``A note on the prediction sum of squares statistic for restricted
  least squares,'' \emph{The American Statistician}, vol.~54, pp. 116--118,
  2000.

\bibitem{Golub1979}
{G. H. Golub, M. Heath and G. Wahba}, ``{Generalized Cross-Validation as a
  Method for Choosing a Good Ridge Parameter},'' \emph{Technometrics}, vol.~21,
  pp. 215--223, 1979.

\bibitem{ObuchiKabashima2016}
{T. Obuchi and Y. Kabashima}, ``{Cross validation in LASSO and its
  acceleration},'' \emph{Journal of Statistical Mechanics: Theory and
  Experiment}, vol. (2016), pp. 053\,304(1--36), 2016.

\bibitem{OpperWinther1996}
{M. Opper and O.~Winther}, ``{A mean field algorithm for Bayes learning in
  large feed-forward neural networks},'' in \emph{Advanced in Neural
  Information Processing Systems 9}, 1997, pp. 225--231.

\bibitem{OpperWinther2001}
{M. Opper and O. Winther}, ``Adaptive and self-averaging
  thouless-anderson-palmer mean-field theory for probabilistic modeling,''
  \emph{Physical Review E}, vol.~64, pp. 056\,131(1--14), 2001.

\bibitem{Winther}
{B. \c{C}akmak, O.~Winther and B. H. Fleury}, ``{S-AMP: Approximate message
  passing for general matrix ensembles},'' in \emph{ITW2014 Proceedings}, 2014,
  pp. 192--196.

\bibitem{Opper}
{M. Opper, B. \c{C}akmak and O. Winther}, ``{A theory of solving TAP equations
  for Ising models with general invariant random matrices},'' \emph{Journal of
  Physics A: Mathematical and Theoretical}, vol.~49, pp. 114\,002(1--24), 2016.

\bibitem{Rangan}
{S. Rangan, A. K. Fletcher, P. Schniter and U. S. Kamilov}, ``{Inference for
  Generalized Linear Models via Alternating Directions and Bethe Free Energy
  Minimization},'' in \emph{ISIT2015 Proceedings}, 2015, pp. 1640--1644.

\bibitem{Krzakala2012}
{F. Krzakala, M. M\'ezard, F. Sausset, Y. Sun and L. Zdeborov\'a},
  ``Probabilistic reconstruction in compressed sensing: algorithms, phase
  diagrams, and threshold achieving matrices,'' \emph{Journal of Statistical
  Mechanics: Theory and Experiment}, vol. 2012, pp. P08\,009 (1--57), 2012.

\bibitem{Nakanishi:15}
{Y. Nakanishi, T. Obuchi, M. Okada and Y. Kabashima}, ``Sparse approximation
  based on a random overcomplete basis,'' \emph{Journal of Statistical
  Mechanics: Theory and Experiment}, vol. (2015), pp. 063\,302(1--30), 2015.

\bibitem{Berkeley}
``{The UC Berkeley Filippenko Group's Supernova Database},''
  http://heracles.astro.berkeley.edu/sndb/.

\bibitem{Silverman:12}
{J.M. Silverman, M. Ganeshalingam, W. Li. and A.V. Filippenko}, ``{Berkeley
  Supernova Ia Program -- III. Spectra near maximum brightness improve the
  accuracy of derived distances to Type Ia supernovae},'' \emph{Mon. Not. R.
  Astron. Soc.}, vol. 425, pp. 1889--1916, 2012.

\bibitem{Uemura:15}
{M. Uemura, K.S. Kawabata, S. Ikeda and K. Maeda}, ``{Variable selection for
  modeling the absolute magnitude at maximum of Type Ia supernovae},''
  \emph{Publ. Astron. Soc. Japan}, vol.~67, pp. 55 (1--9), 2015.

\bibitem{ObuchiKabashimaEUSIPCO2016}
{T. Obuchi and Y. Kabashima}, ``{Sampling approach to sparse approximation
  problem: determining degrees of freedom by simulated annealing},'' in
  \emph{EUSIPCO2016 Proceedings}, 2016, pp. 1247--1251.

\end{thebibliography}



\end{document}